\documentclass[pdflatex,sn-mathphys-num]{sn-jnl}

\usepackage{tabularx}
\usepackage{booktabs}
\usepackage{graphicx}%
\usepackage{multirow}%
\usepackage{amsmath,amssymb,amsfonts}%
\usepackage{amsthm}%
\usepackage{mathrsfs}%
\usepackage[title]{appendix}%
\usepackage{xcolor}%
\usepackage{textcomp}%
\usepackage{manyfoot}%
\usepackage{booktabs}%
\usepackage{algorithm}%
\usepackage{algorithmicx}%
\usepackage{algpseudocode}%
\usepackage{listings}%
\usepackage{longtable}  

\usepackage{subcaption}
\usepackage{arydshln}
\usepackage{booktabs}
\usepackage{xltabular} 
\usepackage{pdflscape} 
\usepackage{geometry}
\usepackage{ragged2e}   
\usepackage{array}      
\newcolumntype{L}{>{\RaggedRight\arraybackslash}X}
\newcolumntype{P}[1]{>{\RaggedRight\arraybackslash}p{#1}}

\theoremstyle{thmstyleone}%
\theoremstyle{thmstyletwo}%

\theoremstyle{thmstylethree}%

\begin{document}


\title[Article Title]{A Diagnostic Framework for AI Agent Behavior}

\author[1]{\fnm{Xichen} \sur{Zhang}}\email{xichen.zhang@smu.ca}

\author*[2]{\fnm{Yingjie} \sur{Zhang}}\email{yingjiezhang@gsm.pku.edu.cn}

\author*[3]{\fnm{Tianshu} \sur{Sun}}\email{tianshusun@ckgsb.edu.cn}

\affil[1]{\orgdiv{Sobey School of Business}, \orgname{Saint Mary's University}, \country{Canada}}

\affil[2]{\orgdiv{Guanghua School of Management}, \orgname{Peking University},  \country{China}}

\affil[3]{\orgdiv{Cheung Kong Graduate School of Business}, \country{China}}

\abstract{
AI agents increasingly act within the same clinical, political, scientific, and social systems that behavioral scientists study. Evaluating these systems requires source-level diagnosis: the same behavioral pattern may arise from an agent's representational substrate or from the roles, objectives, interaction structures, and governance rules that shape its expression. This Perspective proposes a diagnostic framework for AI agent behavior: layer attribution. The foundational computational layer defines what behaviors are possible through architecture, memory, perception, attention, and representation. The behavioral modulation layer shapes how those capacities are expressed through identity, resources, objectives, social interaction, institutional constraints, and governance. The framework clarifies three consequences: surrogate validity is a model-task-layer relation, human-AI divergence provides diagnostic evidence, and governance requires source attribution before intervention. Treating AI agents as behavioral actors therefore requires evaluation methods that determine where behavior originates before deciding how to explain, validate, or govern it.}

\keywords{AI agent behavior, Layer attribution, Behavioral science, Surrogate validity, Human-AI divergence, AI governance}



\maketitle
\begingroup
\renewcommand{\thefootnote}{}
\footnotetext{*Corresponding authors: Yingjie Zhang and Tianshu Sun; Emails:  xichen.zhang@smu.ca, yingjiezhang@gsm.pku.edu.cn, tianshusun@ckgsb.edu.cn}
\addtocounter{footnote}{-1}
\endgroup

\section{Introduction}\label{sec:introduction}

Behavioral scientists increasingly encounter a new kind of actor within the systems they study. AI agents now mediate political persuasion, support clinical decision-making, simulate collective opinion
dynamics, coordinate multi-agent workflows, and contribute to scientific discovery \cite{lin2025persuading,costello2024durably,croxford2025evaluating,wang2025lins,shusterman2025active,gao2025take,chuang2024simulating,li2023camel,hong2023metagpt,wu2024autogen,wang2023scientific}. These systems have moved beyond isolated instruction execution: they interact with people and other agents, act on environments, and feed back into the social systems they inhabit. In many settings, they function as behavioral actors in human systems.

A consistent and troubling pattern has emerged as AI agents enter consequential settings. A clinical AI agent that appears safe in aggregate evaluation may still reflect statistical underrepresentation or uneven performance across patient populations
\cite{croxford2025evaluating,wang2025lins,shusterman2025active}.
Political persuasion systems designed to present information neutrally, or social simulation systems designed to reproduce dynamic human behavior, may still compress opinion heterogeneity or privilege majority narratives \cite{gao2025take,ashery2025emergent,murthy2025one}.
These patterns share a common structure: they recur systematically, often survive additional data or downstream guardrails, and can remain hidden beneath the output surface where much evaluation occurs. In many cases, the relevant source lies below this surface: in what the system has encoded, in what it is optimizing for, or in what its architecture makes possible or
impossible to represent. The result is a systematic gap between what AI agents appear to do under evaluation and what they do when their behavior accumulates, adapts, and shapes real lives in the social systems they inhabit.

Rahwan et al.\ established that AI systems warrant scientific study as behavioral actors with engineering substrates \cite{rahwan2019machine}. That agenda has become urgent
as clinical decision dialogues, multi-agent coordination, and population-scale persuasion have moved from anticipatory to concrete \cite{croxford2025evaluating,wang2025lins,shusterman2025active,li2023camel,hong2023metagpt,wu2024autogen,lin2025persuading,costello2024durably}. Existing machine-behavior, evaluation, alignment, and governance approaches identify consequential AI behaviors, measure its effects, or constrain its expression. They give less guidance on source diagnosis: whether a behavioral pattern arises from the agent's representational substrate or from the roles, objectives, interactions, and institutional rules that shape its
expression. Observing that a clinical AI underperforms across patient populations identifies a problem; layer attribution asks where that problem originates before determining what type of explanation or intervention is appropriate. This Perspective proposes such a framework. The central move is layer attribution: distinguishing whether a behavioral pattern originates from the foundational computational processes that make behavior possible or in the behavioral modulation layer that shapes how it is expressed.

The foundational computational layer, encompassing architecture, memory, perception and attention, and representational structure, defines what behaviors are possible for a given agent.
The behavioral modulation layer, encompassing role assignment, objectives, interaction structures, institutional constraints, and governance, shapes how that potential is expressed in context. The value of this distinction is diagnostic: it asks which layer generated a behavioral pattern and what that source implies for explanation, validation, and intervention. This diagnostic step yields three linked claims. Surrogate validity is a model-task-layer relation, because an AI agent can stand in for humans when the layers relevant to the target behavior are aligned. Human-AI divergence becomes diagnostic evidence, because stable and context-sensitive gaps point to different sources. Governance requires source attribution before intervention, because safer outputs do not by themselves reveal where consequential behavior originates.

\section{Defining AI Agent Behavior}

We define \emph{AI agent behavior} as follows:
\begin{center}
\fbox{
\parbox{0.9\linewidth}{
\textbf{AI Agent Behavior:} The set of context-sensitive, goal-directed actions produced by an AI
agent system as it perceives, reasons about, and acts upon an environment, where those actions
unfold over time, adapt to feedback, and are consequential for the human social systems in which the
agent operates.
}
}
\end{center}

This definition narrows behavior to temporally extended, context-sensitive action. Systems that negotiate with other agents \cite{li2023camel,hong2023metagpt,wu2024autogen}, support clinical
decisions across a patient encounter \cite{croxford2025evaluating,wang2025lins,shusterman2025active},
or durably shift attitudes through sustained, personalized interactions
\cite{costello2024durably,tu2025towards,hao2025generative} exhibit agent behavior in this sense. A single-query response can provide evidence of model behavior, but it becomes agent behavior in the
sense used here when embedded in an unfolding task, interaction, or consequential setting. The criterion is the temporal and relational structure of action, with deployment scale as a secondary consideration: behavior must be understood in context, over time, and in relation to the systems it shapes.

Three properties follow. Agent behavior is \textbf{\textit{interactive}}: bidirectionally coupled with human actors and institutional structures. It is \textbf{\textit{consequential}}: producing effects that accumulate across individuals, groups, and institutions in ways that matter beyond the immediate interaction. It is also \textbf{\textit{attributable}}: traceable, in principle, to the underlying mechanisms that generate it, a theoretical commitment that behavioral science has long made for human behavior and that we
extend here to AI agents. Attributability links a behavioral pattern to a plausible generating source, even when that source remains partially observable. This triad situates AI agent behavior within behavioral science, governance, and engineering evaluation.

\section{A Two-Layer Framework for AI Agent Behavior}

The three properties of AI agent behavior call for a framework that can attribute behavioral patterns to their likely sources. This framework treats human-AI comparison as source diagnosis. It helps researchers ask which source would need to change for an observed behavior to change. We use behavioral science as a selective diagnostic resource. Cognitive science helps distinguish among representation, memory, perception, and attention; social psychology and sociology help distinguish role, norm, interaction, and institutional effects; economics and governance scholarship
help specify how objectives, incentives, and constraints shape action
\cite{nunez2019happened, tsvetkova2024new, rahwan2019machine}. This selective use of behavioral science makes human-AI comparison useful while preserving the difference between AI agency and human
mental life.

\begin{figure}[t]
\centering
\includegraphics[width=1\textwidth]{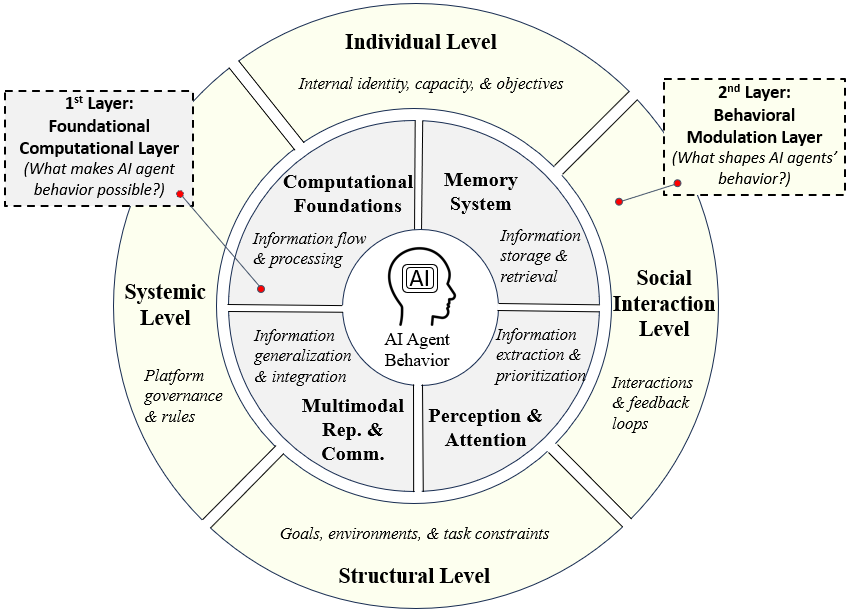}
\caption{Overview of the Two-layer Framework for Modeling AI Agent Behavior}\label{fig:framework-overview}
\end{figure}

We synthesize these resources into a two-layer diagnostic framework for AI agents. The \textbf{foundational computational layer} captures the internal processes that make behavior possible. The \textbf{behavioral modulation layer} captures the conditions that shape how that capacity is expressed across individual, social, structural, and systemic levels. The diagnostic task is to compare behavioral patterns across changes in prompts, roles, objectives,
interaction structures, and governance constraints, then infer whether a pattern is anchored in the agent's substrate, its context of expression, or their interaction. We map each layer onto AI agents
(Figure~\ref{fig:framework-overview}), identifying where behavioral concepts translate directly, where they require reformulation, and where they break down.

\subsection{The Foundational Computational Layer}

\begin{figure}[!htbp]
\centering
\includegraphics[width=1\textwidth]{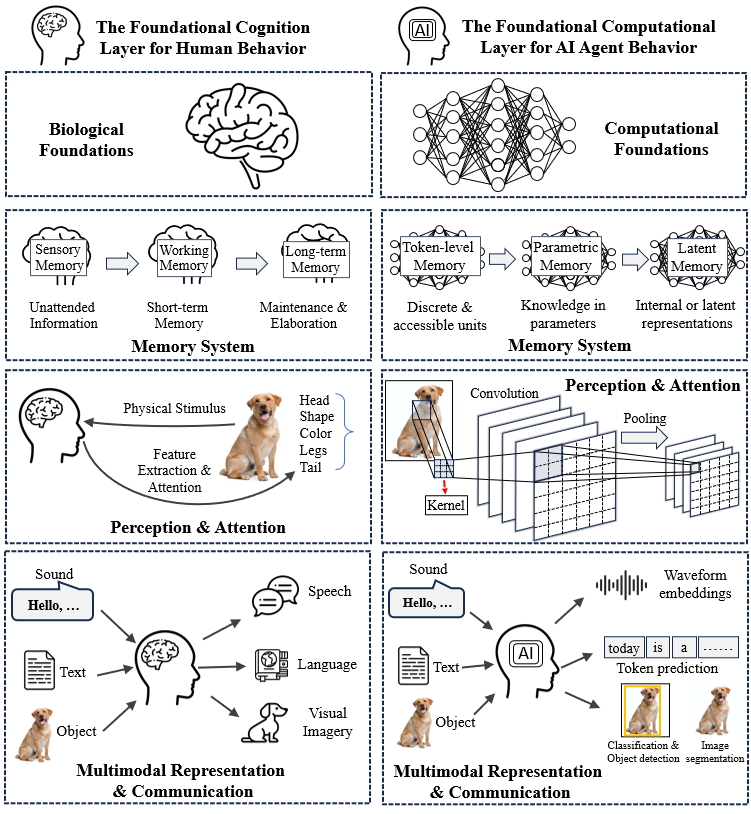}
\caption{Overall comparison of the foundational layer between human and AI agent }\label{fig:comparison-1-layer}
\end{figure}

The foundational computational layer defines what makes AI agent behavior possible. We characterize its technical components by their behavioral significance and their relationship to the analogous process in human cognition. For layer attribution, the central question is whether an observed behavioral pattern remains stable as prompts, roles, objectives, and governance constraints change. Stable patterns suggest limits or tendencies in the agent's representational or architectural possibility space. Overall comparisons between human and AI agents are shown in Figure \ref{fig:comparison-1-layer}.

\textbf{Computational foundations} set the ceiling for what reasoning and action are possible: the architecture determines which input patterns are amplified into behavior and which get suppressed, from which all higher-level agent capabilities emerge \cite{lecun2015deep,
wright2022deep}. \textbf{Memory systems} shape behavior by determining what is accessible, what is treated as settled background knowledge, and what is carried implicitly across an interaction. In AI
agents, memory operates across three levels: \textit{token-level memory} for immediate context, \textit{parametric memory} for learned statistical regularities, and \textit{latent memory} for
implicit internal representations \cite{hu2025memory}, providing a functional analogy to sensory, working, and long-term memory in
human cognition while preserving important differences in mechanism. \textbf{Perception and attention} determine what an agent notices, how information is extracted, and which signals are prioritized,
through learned statistical relations, task prompts, tool access, and context design \cite{zhao2019object, niu2021review, vaswani2017attention}. Finally, \textbf{multimodal representation and communication} enable agents to operate across text, image,
and structured data. The resulting representations remain statistical and indirectly grounded in physical or social experience, constraining how reliably they generalize to novel contexts
\cite{xu2025revealing, xiang2025vision, schulze2025visual}.

Table \ref{tab:human_ai_cognition_comparison} translates these foundational components into diagnostic cues. It links observable behavioral patterns, such as lost context, compressed minority views, misplaced salience, and shallow social interpretation, to the AI substrate that may generate them. The table therefore connects human cognitive anchors to AI-specific mechanisms and clarifies where source-level diagnosis should begin.

\begin{table}[t]
\centering
\footnotesize
\caption{Foundational computational layer: diagnostic cues}
\label{tab:human_ai_cognition_comparison}
\setlength{\tabcolsep}{4pt}
\renewcommand{\arraystretch}{1.15}
\begin{tabularx}{\textwidth}{P{2.1cm} L L L}
\toprule
\textbf{Component} &
\textbf{Human anchor} &
\textbf{AI substrate} &
\textbf{Diagnostic cue} \\
\midrule
\textbf{Foundation}
& Neural adaptation under biological and cognitive constraints \cite{rumelhart1986parallel, hebb2005organization}
& Weighted architectures; training regularities
& Persistent limits in reasoning, planning, abstraction, or generalization may indicate architecture, training, or model class \\

\textbf{Memory system}
& Sensory, working, and long-term memory \cite{atkinson1968human, miller1956magical}
& Context window; parametric and latent memory
& Lost context, recency, stereotyped recall, or compressed minority views may indicate context access, retrieval, or representation \\

\textbf{Perception and attention}
& Perceptual filtering and selective attention \cite{broadbent2013perception, treisman1980feature}
& Pattern recognition; input prioritization
& Misread scenes, missed cues, or misplaced salience may indicate encoding, attention allocation, modality limits, or context design \\

\textbf{Multimodal representation and communication}
& Grounded meaning across modalities \cite{paivio2013imagery, barsalou1999perceptual}
& Cross-modal embedding alignment
& Weak grounding, brittle transfer, or shallow social interpretation may indicate representation or cross-modal alignment \\
\bottomrule
\end{tabularx}
\end{table}

\subsection{The Behavioral Modulation Layer}

\begin{figure}[!htbp]
\centering
\includegraphics[width=1\textwidth]{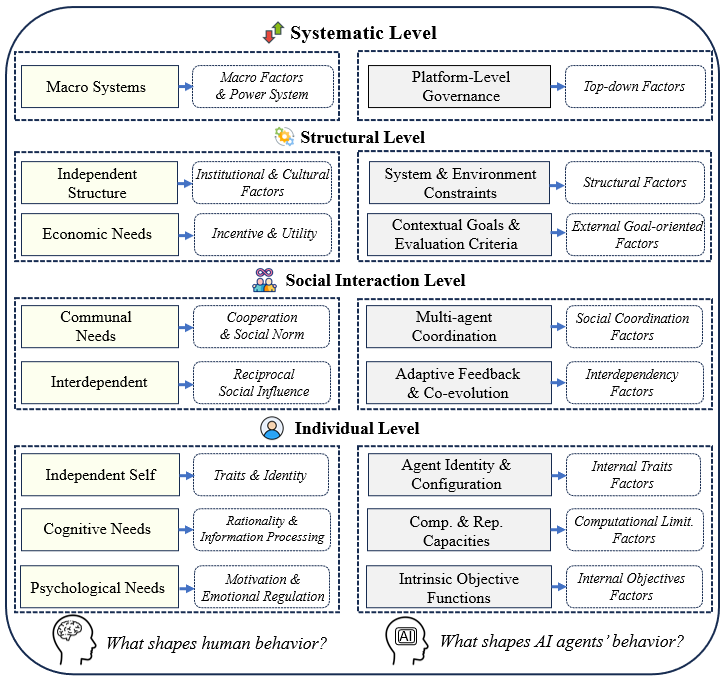}
\caption{Overall comparison of the behavioral modulation layer between human and AI agent }\label{fig:comparison-2-layer}
\end{figure}

The behavioral modulation layer defines \textit{what shapes AI agent behavior} across four hierarchical levels that parallel human behavioral forces: individual, social interaction, structural, and systemic. This layer explains why the same foundational capacity can produce different behaviors under different roles, goals, interaction partners, evaluation criteria, and governance regimes. Translating this structure into AI terms reveals both functional parallels and
key points at which the human analogy diverges. For diagnosis, these levels identify the conditions to perturb when testing whether a behavioral pattern is expression-driven. A pattern that shifts with role, objective, interaction structure, or governance rule points to modulation-layer sources. Figure \ref{fig:comparison-2-layer} illustrates these comparisons across the four levels.

At the \textbf{individual level}, three deployment-facing factors shape what an agent does and how it behaves, given its foundational substrate. \textit{Agent identity and configuration} provide a functional analogue to personality and self-concept in humans: assigned roles, memory states, and goal specifications alter decision-making in ways that are consequential and sometimes introduce systematic biases \cite{abou2025agentic, cao2026biased,
li2026single}. \textit{Resource access and configuration} determine how much of that substrate can be used in a task, including context length, retrieval mechanisms, external tools, memory scaffolds, and planning horizons; these conditions shape whether agents overlook long inputs, overweight recent information, or perform more reliably with structured support \cite{gao2024large, liu2024lost,
zhang2025unpacking, hu2025memory, xu2025llm}. \textit{Optimization and alignment objectives} shape which behaviors are rewarded, discouraged, or selected under task constraints. These objectives
provide a computational analogue to motivation without assuming human psychological needs: an agent optimized for safety may avoid uncertainty in computationally specific ways
\cite{du2025survey, feng2024agile, choudhury2025process}.

At the \textbf{social interaction level}, \textit{multi-agent coordination} produces collective behavior through communication protocols, structured roles, and iterative dialogue, with cooperative
and competitive dynamics emerging from incentive structures in ways that mirror human group behavior while retaining computationally specific dynamics \cite{mordatch2018emergence, hong2023metagpt, wu2024autogen,
abdelnabi2024cooperation, ashery2025emergent}. \textit{Adaptive feedback and co-evolution} captures how agent behavior changes through continuous interactions with environments, other agents, and human feedback, with agents capable of mutual adaptation and self-directed improvement \cite{shinn2023reflexion, renze2024self, liu2025survey, ma2024coevolving}.

At the \textbf{structural level}, \textit{system and environmental constraints} define the boundaries of possible action through safety guardrails, simulated environments, and monitoring
mechanisms, shaping performance outcomes even for highly capable models \cite{xiang2025guardagent,
xie2024osworld, liu2026agentdog}. \textit{Contextual goals and evaluation criteria} determine what agents optimize for in practice, with externally specified objectives varying across domains and
performance metrics shaping priorities under competing demands \cite{abou2025agentic, qi2026towards,
acharya2025agentic}.

At the \textbf{systemic level}, \textit{platform-level governance} encompasses the institutional and regulatory rules that determine how agents are developed, deployed, and governed \cite{raza2025trism}, analogous to the top-down structural forces that shape human behavior across societies. These governance structures define the outermost behavioral envelope within which all lower-level factors operate. Individual-level and structural-level interventions complement this systemic envelope by enforcing safety, transparency, and
accountability within specific deployment contexts
\cite{raza2025trism, shavit2023practices, ferrag2026llm, murugesan2025rise}.

\section{Three Consequences of Layer Attribution}
\label{sec:consequences}

Layer attribution changes the status of AI agents in behavioral science. It asks where similarity or divergence comes from, whether that source remains stable across contexts, and what follows for research and governance. The diagnostic rule is straightforward: patterns that persist across roles, prompts, metrics, or deployment constraints suggest foundational sources; patterns that shift with objectives, roles, interaction structures, or governance rules suggest modulation-layer sources. This rule yields three consequences: surrogate validity becomes a model-task-layer relation, human-AI divergence becomes diagnostic evidence, and governance becomes a problem of source attribution before intervention.

\begin{table}[t]
\centering
\footnotesize
\caption{Three consequences of layer attribution}
\label{tab:three_consequences}
\setlength{\tabcolsep}{4pt}
\renewcommand{\arraystretch}{1.15}
\begin{tabularx}{\textwidth}{P{2.4cm} L L L}
\toprule
\textbf{Consequence} &
\textbf{Diagnostic question} &
\textbf{Layer logic} &
\textbf{Implication} \\
\midrule
\textbf{Surrogate validity}
& Can the agent stand in for humans in this domain?
& Validity requires the relevant foundational process and modulation context to align.
& Surrogate validity is a model-task-layer relation. \\

\textbf{Human-AI divergence}
& What does the difference reveal?
& Stable divergence suggests foundational difference; context-sensitive divergence suggests modulation.
& Divergence becomes diagnostic evidence. \\

\textbf{Governance}
& Where should intervention act?
& Modulation-layer tools shape expression; foundational sources require deeper repair.
& Governance requires source attribution before intervention. \\
\bottomrule
\end{tabularx}
\end{table}

\subsection{Surrogate Validity: When AI Agents Can Stand In for Humans}

Layer attribution reframes surrogate validity as a model-task-layer relation. AI agents function as
human surrogates when the behavior being studied draws on foundational processes and modulation
conditions that are relevantly aligned with the human setting. Where the research question concerns
cognitive processing, heuristic judgment, semantic abstraction, or decision patterns, foundational
alignment may be enough to make AI agents informative. Large language models can reproduce aspects of human reasoning,
decision-making, and cognitive bias across experimental tasks \cite{binz2025foundation,
binz2023using, webb2025brain, chen2025manager, echterhoff2024cognitive, zhuang2025llm,
zhang2025unpacking}, and representation-learning systems can extract salient features and abstract
semantic structure in ways that support human-like categorization or concept use
\cite{xu2025revealing, le2023uncovering, yao2022react}.

Foundational resemblance supports only narrow substitution. Many behavioral questions depend on modulation: identity, role, incentives, social feedback, institutional context, and evaluation
criteria. An agent may reproduce individual-level reasoning biases yet compress opinion heterogeneity in social simulations, especially when alignment procedures or plausibility optimization
suppress minority views or uncertain beliefs \cite{chuang2024simulating, ng2025llm, gao2025take}.
The same model can therefore be a useful surrogate for one behavioral question and a poor surrogate for another. This is the methodological payoff: researchers should specify the target behavior, name
the layer that must align, and validate the agent at that layer.

\subsection{Human-AI Divergence: When Difference Becomes Informative}

Layer attribution also changes how divergence should be interpreted. A difference between human and AI behavior becomes diagnostic evidence when it is read against the layer that produced it. If
divergence persists across prompts, roles, metrics, and deployment constraints, it may reveal a foundational difference in representation, grounding, memory, or reasoning. If divergence changes
when objectives, roles, interaction structures, or evaluation criteria change, it is more likely to reflect modulation-layer expression. The same observed gap therefore implies different scientific uses.

Some gaps should be closed because alignment is the goal. Clinical decision support, domain expertise, and safety-critical coordination often require AI behavior to track human-relevant evidence, institutional standards, and context-specific constraints
\cite{croxford2025evaluating, tu2025towards, swanson2025virtual, zhuge2024agent,
mandal2025evaluating}. Other gaps are useful objects of study. Agents that lacks affect, fatigue, and motivated reasoning can serve as a controlled comparison case for asking which human reasoning patterns survive when biological constraints are removed \cite{binz2023using}. Still other gaps signal boundary conditions. Weak grounded causal models, shallow visual understanding, or poor representation of subjective and cultural experience should limit claims about what AI behavior can reveal in settings that depend on embodied perception, causality, or social meaning \cite{schulze2025visual, pataranutaporn2025simulating,
pataranutaporn2023influencing, batzner2025whose}. In this sense, divergence becomes one of the framework's main data sources.

\subsection{Governance: Intervening at the Right Layer}

The same diagnostic logic gives the framework its governance implication: source attribution should precede intervention choice. Many current alignment and governance tools, including RLHF
\cite{christiano2017deep}, constitutional AI \cite{bai2022constitutional}, and deployment guardrails \cite{luo2025agrail}, operate primarily at the behavioral modulation layer. They shape outputs, incentives, and deployment constraints, and
they can be effective when risks originate at that layer. They provide weaker protection when the source is foundational: biased statistical regularities in parametric memory, weak causal structures,
limited grounding, or poor representation of minority belief structures may continue to shape behavior even when outputs appear compliant \cite{schulze2025visual,
pataranutaporn2025simulating, gao2025take}. The danger is superficial alignment: behavior that satisfies a rule at the output surface while the underlying source of distortion remains intact.

This problem becomes sharper in multi-agent systems. When agents coordinate, adapt, or develop compressed communication protocols, behavior may be generated and transmitted through representations
that are no longer fully legible to human overseers \cite{mordatch2018emergence,
ashery2025emergent}. Output-level monitoring then becomes a weaker governance instrument, because the relevant behavior may lie in interaction patterns, emergent norms, or collective dynamics instead
of isolated responses \cite{abdelnabi2024cooperation}. A governance architecture for AI agents therefore needs more than safer prompts or stricter deployment rules. It needs layer attribution as an operating principle: identify whether the behavioral risks originate in foundational representations, modulation-layer expressions, or their interaction, then choose the intervention layer accordingly. This is where behavioral science can contribute most directly, by turning the
study of behavior into a diagnostic method for AI systems that now act inside human institutions.

\normalsize
\section{Conclusion}

AI agents make a new form of behavioral explanation necessary. Human behavior is usually explained by linking action to cognition, motivation, social context, and institutions. AI agent behavior
requires the same source discipline, translated for engineered systems. A benchmark score, human-likeness rating, or compliant output is insufficient when agents persuade, coordinate,
simulate, and decide inside human institutions.

The next agenda is to make layer attribution operational. Behavioral scientists can design stress tests that vary roles, goals, memory, interaction partners, and governance constraints; AI researchers can identify which representations and architectures make patterns durable; governance scholars can decide whether risks call for output constraints, context redesign, or deeper model repair. The payoff is both conceptual and practical: AI agents can be studied as behavioral actors without mistaking them for humans, and governed as engineered systems without reducing their behavior to isolated outputs.

\bibliography{sn-bibliography}

@article{hu2025memory,
  title={Memory in the age of {AI} agents},
  author={Hu, Yuyang and Liu, Shichun and Yue, Yanwei and Zhang, Guibin and Liu, Boyang and Zhu, Fangyi and Lin, Jiahang and Guo, Honglin and Dou, Shihan and Xi, Zhiheng and others},
  journal={arXiv:2512.13564},
  year={2025}
}

@article{atkinson1968human,
  title={Human memory: A proposed system and its control processes},
  author={Atkinson, Richard C and Shiffrin, Richard M},
  journal={Psychology of Learning and Motivation},
  volume={2},
  pages={89--195},
  year={1968},
  publisher={Elsevier}
}

@article{ferrag2026llm,
  title={{LLM} and {AI} Agents for Autonomous Systems: A Survey of Applications, Datasets, and Security Challenges},
  author={Ferrag, Mohamed Amine and Lakas, Abderrahmane and Tihanyi, Norbert and Debbah, Merouane},
  journal={IEEE Open Journal of Intelligent Transportation Systems},
  volume={7},
  pages={615--657},
  year={2026},
  publisher={IEEE}
}

@article{raza2025trism,
  title={Trism for agentic {AI}: A review of trust, risk, and security management in {LLM}-based agentic multi-agent systems},
  author={Raza, Shaina and Sapkota, Ranjan and Karkee, Manoj and Emmanouilidis, Christos},
  journal={arXiv:2506.04133},
  year={2025}
}

@article{qi2026towards,
  title={Towards transparent and incentive-compatible collaboration in decentralized {LLM} multi-agent systems: A blockchain-driven approach},
  author={Qi, Minfeng and Zhu, Tianqing and Zhang, Lefeng and Li, Ningran and Tan, Yu-an and Zhou, Wanlei},
  journal={IEEE Transactions on Network Science and Engineering},
  year={2026},
  publisher={IEEE}
}

@article{ma2024coevolving,
  title={Coevolving with the other you: Fine-tuning {LLM} with sequential cooperative multi-agent reinforcement learning},
  author={Ma, Hao and Hu, Tianyi and Pu, Zhiqiang and Boyin, Liu and Ai, Xiaolin and Liang, Yanyan and Chen, Min},
  journal={Advances in Neural Information Processing Systems},
  volume={37},
  pages={15497--15525},
  year={2024}
}

@inproceedings{liu2025survey,
  title={A survey on the feedback mechanism of {LLM}-based {AI} agents},
  author={Liu, Zhipeng and Bai, Xuefeng and Chen, Kehai and Chen, Xinyang and Li, Xiucheng and Xiang, Yang and Liu, Jin and Li, Hong-Dong and Wang, Yaowei and Nie, Liqiang and others},
  booktitle={Proceedings of the Thirty-Fourth International Joint Conference on Artificial Intelligence},
  pages={10582--10592},
  year={2025}
}

@article{renze2024self,
  title={Self-reflection in {LLM} agents: Effects on problem-solving performance},
  author={Renze, Matthew and Guven, Erhan},
  journal={arXiv:2405.06682},
  year={2024}
}

@article{du2025survey,
  title={A survey on the optimization of large language model-based agents},
  author={Du, Shangheng and Zhao, Jiabao and Shi, Jinxin and Xie, Zhentao and Jiang, Xin and Bai, Yanhong and He, Liang},
  journal={ACM Computing Surveys},
  year={2025},
  publisher={ACM New York, NY}
}

@article{gao2025take,
  title={Take caution in using {LLMs} as human surrogates},
  author={Gao, Yuan and Lee, Dokyun and Burtch, Gordon and Fazelpour, Sina},
  journal={Proceedings of the National Academy of Sciences},
  volume={122},
  number={24},
  pages={e2501660122},
  year={2025},
  publisher={National Academy of Sciences}
}

@article{gao2024large,
  title={Large language models empowered agent-based modeling and simulation: A survey and perspectives},
  author={Gao, Chen and Lan, Xiaochong and Li, Nian and Yuan, Yuan and Ding, Jingtao and Zhou, Zhilun and Xu, Fengli and Li, Yong},
  journal={Humanities and Social Sciences Communications},
  volume={11},
  number={1},
  pages={1--24},
  year={2024},
  publisher={Palgrave}
}

@article{abou2025agentic,
  title={Agentic {AI}: a comprehensive survey of architectures, applications, and future directions},
  author={Abou Ali, Mohamad and Dornaika, Fadi and Charafeddine, Jinan},
  journal={Artificial Intelligence Review},
  volume={59},
  number={1},
  pages={11},
  year={2025},
  publisher={Springer}
}

@article{xiang2025vision,
  title={A vision--language foundation model for precision oncology},
  author={Xiang, Jinxi and Wang, Xiyue and Zhang, Xiaoming and Xi, Yinghua and Eweje, Feyisope and Chen, Yijiang and Li, Yuchen and Bergstrom, Colin and Gopaulchan, Matthew and Kim, Ted and others},
  journal={Nature},
  volume={638},
  number={8051},
  pages={769--778},
  year={2025},
  publisher={Nature Publishing Group UK London}
}

@article{vaswani2017attention,
  title={Attention is all you need},
  author={Vaswani, Ashish and Shazeer, Noam and Parmar, Niki and Uszkoreit, Jakob and Jones, Llion and Gomez, Aidan N and Kaiser, {\L}ukasz and Polosukhin, Illia},
  journal={Advances in Neural Information Processing Systems},
  volume={30},
  year={2017}
}

@article{niu2021review,
  title={A review on the attention mechanism of deep learning},
  author={Niu, Zhaoyang and Zhong, Guoqiang and Yu, Hui},
  journal={Neurocomputing},
  volume={452},
  pages={48--62},
  year={2021},
  publisher={Elsevier}
}

@article{zhao2019object,
  title={Object detection with deep learning: A review},
  author={Zhao, Zhong-Qiu and Zheng, Peng and Xu, Shou-tao and Wu, Xindong},
  journal={IEEE Transactions on Neural nNetworks and lLearning Systems},
  volume={30},
  number={11},
  pages={3212--3232},
  year={2019},
  publisher={IEEE}
}

@article{wright2022deep,
  title={Deep physical neural networks trained with backpropagation},
  author={Wright, Logan G and Onodera, Tatsuhiro and Stein, Martin M and Wang, Tianyu and Schachter, Darren T and Hu, Zoey and McMahon, Peter L},
  journal={Nature},
  volume={601},
  number={7894},
  pages={549--555},
  year={2022},
  publisher={Nature Publishing Group UK London}
}

@article{lecun2015deep,
  title={Deep learning},
  author={LeCun, Yann and Bengio, Yoshua and Hinton, Geoffrey},
  journal={Nature},
  volume={521},
  number={7553},
  pages={436--444},
  year={2015},
  publisher={Nature Publishing Group UK London}
}

@article{wang2023scientific,
  title={Scientific discovery in the age of artificial intelligence},
  author={Wang, Hanchen and Fu, Tianfan and Du, Yuanqi and Gao, Wenhao and Huang, Kexin and Liu, Ziming and Chandak, Payal and Liu, Shengchao and Van Katwyk, Peter and Deac, Andreea and others},
  journal={Nature},
  volume={620},
  number={7972},
  pages={47--60},
  year={2023},
  publisher={Nature Publishing Group UK London}
}

@article{binz2025foundation,
  title={A foundation model to predict and capture human cognition},
  author={Binz, Marcel and Akata, Elif and Bethge, Matthias and Br{\"a}ndle, Franziska and Callaway, Fred and Coda-Forno, Julian and Dayan, Peter and Demircan, Can and Eckstein, Maria K and {\'E}ltet{\H{o}}, No{\'e}mi and others},
  journal={Nature},
  pages={1--8},
  year={2025},
  publisher={Nature Publishing Group UK London}
}

@inproceedings{echterhoff2024cognitive,
  title={Cognitive bias in decision-making with {LLMs}},
  author={Echterhoff, Jessica Maria and Liu, Yao and Alessa, Abeer and McAuley, Julian and He, Zexue},
  booktitle={Findings of the association for computational linguistics: EMNLP 2024},
  pages={12640--12653},
  year={2024}
}

@inproceedings{zhuang2025llm,
  title={{LLM} Agents Can Be Choice-Supportive Biased Evaluators: An Empirical Study},
  author={Zhuang, Nan and Cao, Boyu and Yang, Yi and Xu, Jing and Xu, Mingda and Wang, Yuxiao and Liu, Qi},
  booktitle={Proceedings of the AAAI Conference on Artificial Intelligence},
  volume={39},
  number={25},
  pages={26436--26444},
  year={2025}
}

@inproceedings{chuang2024simulating,
  title={Simulating opinion dynamics with networks of {LLM-based} agents},
  author={Chuang, Yun-Shiuan and Goyal, Agam and Harlalka, Nikunj and Suresh, Siddharth and Hawkins, Robert and Yang, Sijia and Shah, Dhavan and Hu, Junjie and Rogers, Timothy},
  booktitle={Findings of the association for computational linguistics: NAACL 2024},
  pages={3326--3346},
  year={2024}
}

@inproceedings{batzner2025whose,
  title={Whose Personae? {Synthetic} Persona Experiments in {LLM} Research and Pathways to Transparency},
  author={Batzner, Jan and Stocker, Volker and Tang, Bingjun and Natarajan, Anusha and Chen, Qinhao and Schmid, Stefan and Kasneci, Gjergji},
  booktitle={Proceedings of the AAAI/ACM Conference on AI, Ethics, and Society},
  volume={8},
  number={1},
  pages={343--354},
  year={2025}
}

@article{zhuge2024agent,
  title={{Agent-as-a-Judge}: Evaluate agents with agents},
  author={Zhuge, Mingchen and Zhao, Changsheng and Ashley, Dylan and Wang, Wenyi and Khizbullin, Dmitrii and Xiong, Yunyang and Liu, Zechun and Chang, Ernie and Krishnamoorthi, Raghuraman and Tian, Yuandong and others},
  journal={arXiv:2410.10934},
  year={2024}
}

@inproceedings{ng2025llm,
  title={Are {LLM}-Powered Social Media Bots Realistic?},
  author={Ng, Lynnette Hui Xian and Carley, Kathleen M},
  booktitle={International Conference on Social Computing, Behavioral-Cultural Modeling and Prediction and Behavior Representation in Modeling and Simulation},
  pages={14--23},
  year={2025},
  organization={Springer}
}

@article{xu2025llm,
  title={{LLM}-Based Agents for Tool Learning: A Survey},
  author={Xu, Weikai and Huang, Chengrui and Gao, Shen and Shang, Shuo},
  journal={Data Science and Engineering},
  pages={1--31},
  year={2025},
  publisher={Springer}
}

@article{zhang2025unpacking,
  title={Unpacking the Decision Logic of {LLM}-based Agents: Evidence from the Newsvendor Problem},
  author={Zhang, Xichen and Cavusoglu, Hasan},
  journal={Available at SSRN 5337019},
  year={2025}
}

@article{shinn2023reflexion,
  title={Reflexion: Language agents with verbal reinforcement learning},
  author={Shinn, Noah and Cassano, Federico and Gopinath, Ashwin and Narasimhan, Karthik and Yao, Shunyu},
  journal={Advances in Neural Information Processing Systems},
  volume={36},
  pages={8634--8652},
  year={2023}
}

@book{broadbent2013perception,
  title={Perception and communication},
  author={Broadbent, Donald Eric},
  year={2013},
  publisher={Elsevier},
  address = {New York}
}

@article{treisman1980feature,
  title={A feature-integration theory of attention},
  author={Treisman, Anne M and Gelade, Garry},
  journal={Cognitive Psychology},
  volume={12},
  number={1},
  pages={97--136},
  year={1980},
  publisher={Elsevier}
}

@article{barsalou1999perceptual,
  title={Perceptual symbol systems},
  author={Barsalou, Lawrence W},
  journal={Behavioral and Brain Sciences},
  volume={22},
  number={4},
  pages={577--660},
  year={1999},
  publisher={Cambridge University Press}
}

@inproceedings{murthy2025one,
  title={One fish, two fish, but not the whole sea: Alignment reduces language models’ conceptual diversity},
  author={Murthy, Sonia Krishna and Ullman, Tomer and Hu, Jennifer},
  booktitle={Proceedings of the 2025 Conference of the Nations of the Americas Chapter of the Association for Computational Linguistics: Human Language Technologies (Volume 1: Long Papers)},
  pages={11241--11258},
  year={2025}
}

@book{paivio2013imagery,
  title={Imagery and verbal processes},
  author={Paivio, Allan},
  year={2013},
  publisher={Psychology Press}, 
  address = {London}
}

@article{miller1956magical,
  title={The magical number seven, plus or minus two: Some limits on our capacity for processing information.},
  author={Miller, George A},
  journal={Psychological Review},
  volume={63},
  number={2},
  pages={81},
  year={1956},
  publisher={American Psychological Association}
}

@book{hebb2005organization,
  title={The organization of behavior: A neuropsychological theory},
  author={Hebb, Donald Olding},
  year={2005},
  publisher={Psychology Press},
  address = {London}
}

@book{rumelhart1986parallel,
  title={Parallel distributed processing, volume 1: Explorations in the microstructure of cognition: Foundations},
  author={Rumelhart, David E and McClelland, James L and PDP Research Group and others},
  year={1986},
  publisher={MIT press},
  address = {Cambridge}
}

@inproceedings{luo2025agrail,
  title={Agrail: A lifelong agent guardrail with effective and adaptive safety detection},
  author={Luo, Weidi and Dai, Shenghong and Liu, Xiaogeng and Banerjee, Suman and Sun, Huan and Chen, Muhao and Xiao, Chaowei},
  booktitle={Proceedings of the 63rd Annual Meeting of the Association for Computational Linguistics (Volume 1: Long Papers)},
  pages={8104--8139},
  year={2025}
}

@article{bai2022constitutional,
  title={Constitutional {AI}: Harmlessness from {AI} feedback},
  author={Bai, Yuntao and Kadavath, Saurav and Kundu, Sandipan and Askell, Amanda and Kernion, Jackson and Jones, Andy and Chen, Anna and Goldie, Anna and Mirhoseini, Azalia and McKinnon, Cameron and others},
  journal={arXiv:2212.08073},
  year={2022}
}

@article{yao2022react,
  title={{ReAct}: Synergizing reasoning and acting in language models},
  author={Yao, Shunyu and Zhao, Jeffrey and Yu, Dian and Du, Nan and Shafran, Izhak and Narasimhan, Karthik and Cao, Yuan},
  journal={arXiv:2210.03629},
  year={2022}
}

@article{christiano2017deep,
  title={Deep reinforcement learning from human preferences},
  author={Christiano, Paul F and Leike, Jan and Brown, Tom and Martic, Miljan and Legg, Shane and Amodei, Dario},
  journal={Advances in Neural Information Processing Systems},
  volume={30},
  year={2017}
}

@article{murugesan2025rise,
  title={The rise of agentic {AI}: implications, concerns, and the path forward},
  author={Murugesan, San},
  journal={IEEE Intelligent Systems},
  volume={40},
  number={2},
  pages={8--14},
  year={2025},
  publisher={IEEE}
}

@article{shavit2023practices,
  title={Practices for governing agentic {AI} systems},
  author={Shavit, Yonadav and Agarwal, Sandhini and Brundage, Miles and Adler, Steven and O’Keefe, Cullen and Campbell, Rosie and Lee, Teddy and Mishkin, Pamela and Eloundou, Tyna and Hickey, Alan and others},
  journal={Research Paper, OpenAI},
  year={2023}
}

@article{acharya2025agentic,
  title={Agentic {AI}: Autonomous intelligence for complex goals: A comprehensive survey},
  author={Acharya, Deepak Bhaskar and Kuppan, Karthigeyan and Divya, B},
  journal={IEEE Access},
  volume={13},
  pages={18912--18936},
  year={2025},
  publisher={IEEE}
}

@article{abdelnabi2024cooperation,
  title={Cooperation, competition, and maliciousness: {LLM}-stakeholders interactive negotiation},
  author={Abdelnabi, Sahar and Gomaa, Amr and Sivaprasad, Sarath and Sch{\"o}nherr, Lea and Fritz, Mario},
  journal={Advances in Neural Information Processing Systems},
  volume={37},
  pages={83548--83599},
  year={2024}
}

@inproceedings{wu2024autogen,
  title={Autogen: Enabling next-gen {LLM} applications via multi-agent conversations},
  author={Wu, Qingyun and Bansal, Gagan and Zhang, Jieyu and Wu, Yiran and Li, Beibin and Zhu, Erkang and Jiang, Li and Zhang, Xiaoyun and Zhang, Shaokun and Liu, Jiale and others},
  booktitle={First conference on language modeling},
  year={2024}
}

@inproceedings{hong2023metagpt,
  title={{MetaGPT}: Meta programming for a multi-agent collaborative framework},
  author={Hong, Sirui and Zhuge, Mingchen and Chen, Jonathan and Zheng, Xiawu and Cheng, Yuheng and Wang, Jinlin and Zhang, Ceyao and Wang, Zili and Yau, Steven Ka Shing and Lin, Zijuan and others},
  booktitle={The 12th International Conference on Learning Representations},
  pages={23247--23275},
  year={2023}
}

@inproceedings{mordatch2018emergence,
  title={Emergence of grounded compositional language in multi-agent populations},
  author={Mordatch, Igor and Abbeel, Pieter},
  booktitle={Proceedings of the AAAI conference on artificial intelligence},
  volume={32},
  number={1},
  year={2018}
}

@article{choudhury2025process,
  title={Process reward models for {LLM} agents: Practical framework and directions},
  author={Choudhury, Sanjiban},
  journal={arXiv:2502.10325},
  year={2025}
}

@article{feng2024agile,
  title={Agile: A novel reinforcement learning framework of {LLM} agents},
  author={Feng, Peiyuan and He, Yichen and Huang, Guanhua and Lin, Yuan and Zhang, Hanchong and Zhang, Yuchen and Li, Hang},
  journal={Advances in Neural Information Processing Systems},
  volume={37},
  pages={5244--5284},
  year={2024}
}

@article{costello2024durably,
  title={Durably reducing conspiracy beliefs through dialogues with {AI}},
  author={Costello, Thomas H and Pennycook, Gordon and Rand, David G},
  journal={Science},
  volume={385},
  pages={eadq1814},
  year={2024},
  publisher={American Association for the Advancement of Science}
}

@article{ashery2025emergent,
  title={Emergent social conventions and collective bias in {LLM} populations},
  author={Ashery, Ariel Flint and Aiello, Luca Maria and Baronchelli, Andrea},
  journal={Science Advances},
  volume={11},
  number={20},
  pages={eadu9368},
  year={2025},
  publisher={American Association for the Advancement of Science}
}

@article{hao2025generative,
  title={How generative {AI} shapes user perceived value and adoption intention in digital museum experiences},
  author={Hao, Xiaoyan and Xu, Junping and Wang, Ying},
  journal={npj Heritage Science},
  volume={13},
  number={1},
  pages={608},
  year={2025},
  publisher={Springer International Publishing Cham}
}

@article{lin2025persuading,
  title={Persuading voters using human--artificial intelligence dialogues},
  author={Lin, Hause and Czarnek, Gabriela and Lewis, Benjamin and White, Joshua P and Berinsky, Adam J and Costello, Thomas and Pennycook, Gordon and Rand, David G},
  journal={Nature},
  pages={1--8},
  year={2025},
  publisher={Nature Publishing Group UK London}
}

@article{webb2025brain,
  title={A brain-inspired agentic architecture to improve planning with {LLMs}},
  author={Webb, Taylor and Mondal, Shanka Subhra and Momennejad, Ida},
  journal={Nature Communications},
  volume={16},
  number={1},
  pages={8633},
  year={2025},
  publisher={Nature Publishing Group UK London}
}

@article{tu2025towards,
  title={Towards conversational diagnostic artificial intelligence},
  author={Tu, Tao and Schaekermann, Mike and Palepu, Anil and Saab, Khaled and Freyberg, Jan and Tanno, Ryutaro and Wang, Amy and Li, Brenna and Amin, Mohamed and Cheng, Yong and others},
  journal={Nature},
  pages={1--9},
  year={2025},
  publisher={Nature Publishing Group UK London}
}

@article{mandal2025evaluating,
  title={Evaluating large language model agents for automation of atomic force microscopy},
  author={Mandal, Indrajeet and Soni, Jitendra and Zaki, Mohd and Smedskjaer, Morten M and Wondraczek, Katrin and Wondraczek, Lothar and Gosvami, Nitya Nand and Krishnan, NM Anoop},
  journal={Nature Communications},
  volume={16},
  number={1},
  pages={9104},
  year={2025},
  publisher={Nature Publishing Group UK London}
}

@article{shusterman2025active,
  title={An active inference strategy for prompting reliable responses from large language models in medical practice},
  author={Shusterman, Roma and Waters, Allison C and O’Neill, Shannon and Bangs, Marshall and Luu, Phan and Tucker, Don M},
  journal={npj Digital Medicine},
  volume={8},
  number={1},
  pages={119},
  year={2025},
  publisher={Nature Publishing Group UK London}
}

@article{swanson2025virtual,
  title={The Virtual Lab of {AI} agents designs new {SARS-CoV-2} nanobodies},
  author={Swanson, Kyle and Wu, Wesley and Bulaong, Nash L and Pak, John E and Zou, James},
  journal={Nature},
  volume={646},
  number={8085},
  pages={716--723},
  year={2025},
  publisher={Nature Publishing Group UK London}
}

@article{wang2025lins,
  title={LINS: A general medical Q\&A framework for enhancing the quality and credibility of {LLM}-generated responses},
  author={Wang, Sheng and Zhao, Fangyuan and Bu, Dechao and Lu, Yunwei and Gong, Ming and Liu, Hongjie and Yang, Zhaohui and Zeng, Xiaoxi and Yuan, Zhiyuan and Wan, Baoping and others},
  journal={Nature Communications},
  volume={16},
  number={1},
  pages={9076},
  year={2025},
  publisher={Nature Publishing Group UK London}
}

@article{croxford2025evaluating,
  title={Evaluating clinical {AI} summaries with large language models as judges},
  author={Croxford, Emma and Gao, Yanjun and First, Elliot and Pellegrino, Nicholas and Schnier, Miranda and Caskey, John and Oguss, Madeline and Wills, Graham and Chen, Guanhua and Dligach, Dmitriy and others},
  journal={npj Digital Medicine},
  volume={8},
  number={1},
  pages={640},
  year={2025},
  publisher={Nature Publishing Group UK London}
}

@article{pataranutaporn2023influencing,
  title={Influencing human-{AI} interaction by priming beliefs about {AI} can increase perceived trustworthiness, empathy and effectiveness},
  author={Pataranutaporn, Pat and Liu, Ruby and Finn, Ed and Maes, Pattie},
  journal={Nature Machine Intelligence},
  volume={5},
  number={10},
  pages={1076--1086},
  year={2023},
  publisher={Nature Publishing Group UK London}
}

@article{xu2025revealing,
  title={Revealing emergent human-like conceptual representations from language prediction},
  author={Xu, Ningyu and Zhang, Qi and Du, Chao and Luo, Qiang and Qiu, Xipeng and Huang, Xuanjing and Zhang, Menghan},
  journal={Proceedings of the National Academy of Sciences},
  volume={122},
  number={44},
  pages={e2512514122},
  year={2025},
  publisher={National Academy of Sciences}
}

@article{pataranutaporn2025simulating,
  title={Simulating human well-being with large language models: Systematic validation and misestimation across 64,000 individuals from 64 countries},
  author={Pataranutaporn, Pat and Powdthavee, Nattavudh and Archiwaranguprok, Chayapatr and Maes, Pattie},
  journal={Proceedings of the National Academy of Sciences},
  volume={122},
  number={48},
  pages={e2519394122},
  year={2025},
  publisher={National Academy of Sciences}
}

@article{le2023uncovering,
  title={Uncovering the semantics of concepts using {GPT}-4},
  author={Le Mens, Ga{\"e}l and Kov{\'a}cs, Bal{\'a}zs and Hannan, Michael T and Pros, Guillem},
  journal={Proceedings of the National Academy of Sciences},
  volume={120},
  number={49},
  pages={e2309350120},
  year={2023},
  publisher={National Academy of Sciences}
}

@article{binz2023using,
  title={Using cognitive psychology to understand {GPT}-3},
  author={Binz, Marcel and Schulz, Eric},
  journal={Proceedings of the National Academy of Sciences},
  volume={120},
  number={6},
  pages={e2218523120},
  year={2023},
  publisher={National Academy of Sciences}
}

@article{chen2025manager,
  title={A manager and an {AI} walk into a bar: does {ChatGPT} make biased decisions like we do?},
  author={Chen, Yang and Kirshner, Samuel N and Ovchinnikov, Anton and Andiappan, Meena and Jenkin, Tracy},
  journal={Manufacturing \& Service Operations Management},
  volume={27},
  number={2},
  pages={354--368},
  year={2025},
  publisher={INFORMS}
}

@inproceedings{li2026single,
  title={From Single to Societal: Analyzing Persona-Induced Bias in Multi-Agent Interactions},
  author={Li, Jiayi and Liu, Xiao and Feng, Yansong},
  booktitle={Proceedings of the AAAI Conference on Artificial Intelligence},
  volume={40},
  number={37},
  pages={31609--31617},
  year={2026}
}

@article{liu2024lost,
  title={Lost in the middle: How language models use long contexts},
  author={Liu, Nelson F and Lin, Kevin and Hewitt, John and Paranjape, Ashwin and Bevilacqua, Michele and Petroni, Fabio and Liang, Percy},
  journal={Transactions of the Association for Computational Linguistics},
  volume={12},
  pages={157--173},
  year={2024}
}

@article{cao2026biased,
  title={From Biased Chatbots to Biased Agents: Examining Role Assignment Effects on {LLM} Agent Robustness},
  author={Cao, Linbo and Sun, Lihao and Yue, Yang},
  journal={arXiv:2602.12285},
  year={2026}
}

@article{schulze2025visual,
  title={Visual cognition in multimodal large language models},
  author={Schulze Buschoff, Luca M and Akata, Elif and Bethge, Matthias and Schulz, Eric},
  journal={Nature Machine Intelligence},
  volume={7},
  number={1},
  pages={96--106},
  year={2025},
  publisher={Nature Publishing Group UK London}
}

@article{li2023camel,
  title={Camel: Communicative agents for ``mind" exploration of large language model society},
  author={Li, Guohao and Hammoud, Hasan and Itani, Hani and Khizbullin, Dmitrii and Ghanem, Bernard},
  journal={Advances in Neural Information Processing Systems},
  volume={36},
  pages={51991--52008},
  year={2023}
}

@article{rahwan2019machine,
  title={Machine behaviour},
  author={Rahwan, Iyad and Cebrian, Manuel and Obradovich, Nick and Bongard, Josh and Bonnefon, Jean-Fran{\c{c}}ois and Breazeal, Cynthia and Crandall, Jacob W and Christakis, Nicholas A and Couzin, Iain D and Jackson, Matthew O and others},
  journal={Nature},
  volume={568},
  number={7753},
  pages={477--486},
  year={2019},
  publisher={Nature Publishing Group UK London}
}

@article{tsvetkova2024new,
  title={A new sociology of humans and machines},
  author={Tsvetkova, Milena and Yasseri, Taha and Pescetelli, Niccolo and Werner, Tobias},
  journal={Nature Human Behaviour},
  volume={8},
  number={10},
  pages={1864--1876},
  year={2024},
  publisher={Nature Publishing Group UK London}
}

@article{nunez2019happened,
  title={What happened to cognitive science?},
  author={N{\'u}{\~n}ez, Rafael and Allen, Michael and Gao, Richard and Miller Rigoli, Carson and Relaford-Doyle, Josephine and Semenuks, Arturs},
  journal={Nature Human Behaviour},
  volume={3},
  number={8},
  pages={782--791},
  year={2019},
  publisher={Nature Publishing Group UK London}
}

@article{liu2026agentdog,
  title={{AgentDog}: A Diagnostic Guardrail Framework for {AI} Agent Safety and Security},
  author={Liu, Dongrui and Ren, Qihan and Qian, Chen and Shao, Shuai and Xie, Yuejin and Li, Yu and Yang, Zhonghao and Luo, Haoyu and Wang, Peng and Liu, Qingyu and others},
  journal={arXiv:2601.18491},
  year={2026}
}

@article{xie2024osworld,
  title={OSworld: Benchmarking multimodal agents for open-ended tasks in real computer environments},
  author={Xie, Tianbao and Zhang, Danyang and Chen, Jixuan and Li, Xiaochuan and Zhao, Siheng and Cao, Ruisheng and Hua, Toh J and Cheng, Zhoujun and Shin, Dongchan and Lei, Fangyu and others},
  journal={Advances in Neural Information Processing Systems},
  volume={37},
  pages={52040--52094},
  year={2024}
}

@inproceedings{xiang2025guardagent,
  title={{GuardAgent}: Safeguard {LLM} agents via knowledge-enabled reasoning},
  author={Xiang, Zhen and Zheng, Linzhi and Li, Yanjie and Hong, Junyuan and Li, Qinbin and Xie, Han and Zhang, Jiawei and Xiong, Zidi and Xie, Chulin and Bastian, Nathaniel D and others},
  booktitle={ICML 2025 workshop on computer use agents},
  year={2025}
}

\end{document}